\documentclass[10pt,twocolumn,letterpaper]{article}
\usepackage{cvpr}

\definecolor{cvprblue}{rgb}{0.21,0.49,0.74}
\usepackage[pagebackref,breaklinks,colorlinks,allcolors=cvprblue]{hyperref}
\usepackage{dblfloatfix}
\usepackage{graphicx} 
\usepackage{float}
\usepackage{balance}
\def\paperID{28} 
\def\confName{CVPR}
\def\confYear{2025}

\title{Vectorized Region Based Brush Strokes for Artistic Rendering}

\author{Jeripothula Prudviraj\\
TCS Research\\
{\tt\small prudviraj.jeripothula@tcs.com}
\and
Vikram Jamwal\\
TCS Research\\
{\tt\small vikram.jamwal@tcs.com}
}

\newcommand{\Bezier}{B\'{e}zier}
\begin{document}

\maketitle

\begin{figure*}
  \includegraphics[width=1.0 \textwidth]{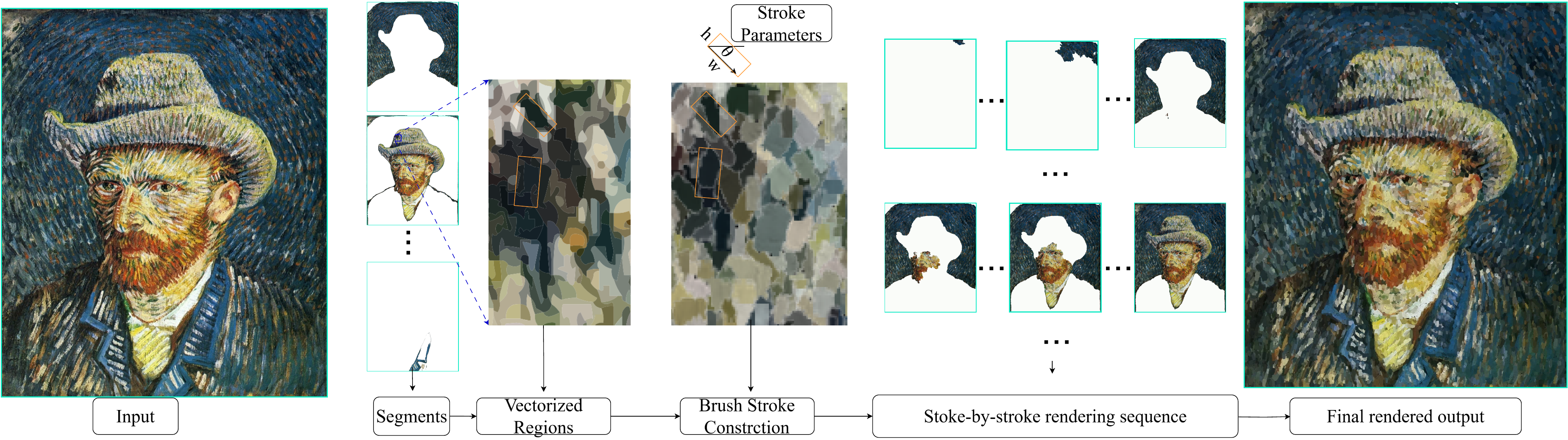}
  \caption{Our painting process involves (i) dividing the input image into a set of pragmatic regions using segmentation and vectorization, (ii) estimating the stroke parameters for each segment-specific vectorized region by converting it into minimum rotated rectangular polygons, (iii) transforming a template brush based on the estimated polygon parameters and perform blending, and finally (iv) sequentially rendering the brush strokes. This process fills the image stroke by stroke, region by region, and segment by segment, following a hierarchical fashion.} 
  \label{fig:teaser}
\end{figure*}

\begin{abstract}
Creating a stroke-by-stroke evolution process of a visual artwork tries to bridge the emotional and educational gap between the finished static artwork and its creation process. 
Recent stroke-based painting systems focus on capturing stroke details by predicting and iteratively refining stroke parameters to maximize the similarity between the input image and the rendered output. However, these methods often struggle to produce stroke compositions that align with artistic principles and intent. To address this, we explore an image-to-painting method that (i) facilitates semantic guidance for brush strokes in targeted regions, (ii) computes the brush stroke parameters, and (iii) establishes a sequence among segments and strokes to sequentially render the final painting. Experimental results on various input image types, such as face images, paintings, and photographic images, show that our method aligns with a region-based painting strategy while rendering a painting with high fidelity and superior stroke quality.
\end{abstract}

\section{Introduction}

Visual art is a fundamental form of human expression, allowing us to explore, communicate, and preserve ideas, emotions, and experiences. However, artworks are typically encountered only as static images. This presentation often omits the dynamic, interactive qualities that characterize the process of artistic creation. Imagine witnessing a painting unfold in real time—each brushstroke gradually appearing, each region taking form, transforming an empty canvas into a finished masterpiece. This dynamic perspective not only enriches creative and emotional engagement but also bridges educational gaps between finished artworks and the process of their creation.

Our work focuses on stroke-based rendering (SBR) that simulates artwork evolution, capturing the essence of artistic creation stroke by stroke. This enables users to engage with how art emerges, offering insights into technique and style, and fostering immersive experiences across domains such as collaborative co-drawing, art education, and virtual exhibitions. By simulating the evolution of art in diverse scenarios, we explore how art is created, learned, and experienced.

Recently, automatic painting systems have gained traction for their utility in digital art, education, and creative applications~\cite{sylaiou2017exploring, hu2024towards, song2024processpainter, nolte2022stroke}. Painting involves intricate steps reflecting the artist’s intent and approach. Artists often employ strategies based on both \textit{how to paint} and \textit{where to paint}. Some start with outlines, others paint region by region—e.g., background, then face, hat, shirt—producing semantically meaningful sequences as noted in~\cite{de2024segmentation}. These choices affect both composition and interpretation.

Most existing SBR approaches~\cite{liu2021paint, nolte2022stroke, tong2022im2oil, zou2021stylized} focus on decomposing input images into vectorized brushstroke sequences defined by shape, position, and color. These are rendered via rule-based~\cite{zeng2009image}, neural-based~\cite{huang2019learning, zou2021stylized, singh2022intelli}, or simulation-based~\cite{chen2015wetbrush} methods. However, many of these lack hierarchical semantic understanding~\cite{hu2024towards} and optimize only pixel-level loss~\cite{de2024segmentation}, leading to rendering processes that diverge from real-world artistic workflows~\cite{chamberlain2016genesis, cohen2005look, glazek2012visual, perdreau2015drawing, tchalenko2009segmentation}. Moreover, such models are often computationally intensive and unsuitable for high-resolution images.

Some recent studies~\cite{hu2024towards, de2024segmentation, singh2022intelli, hu2023stroke} address semantic-region-driven SBR. Intelli-Paint~\cite{singh2022intelli} uses object detection and sliding attention but lacks hierarchical semantics. Guevara~\textit{et al.}~\cite{de2024segmentation} applied DETR~\cite{carion2020end} and attention to predict render regions across grid patches. Hu~\textit{et al.}~\cite{hu2024towards} used multi-granular alignment with segmentation~\cite{kirillov2023segment} and depth maps~\cite{yang2024depth}. DPPR~\cite{hu2023stroke} applies reinforcement learning to region prediction, but its regions are uniform and misaligned with painting semantics. Limitations of these methods include: (i) an overemphasis on image reconstruction via pixel-level loss; (ii) stroke generation lacking realistic sequencing; and (iii) neglecting the structured nature of artistic workflows.

To address these issues, we propose a region-driven brush stroke approach for SBR. Our method combines segmentation and SVG vectorization to determine both \textit{where to paint} and \textit{what to paint}. Segmentation provides object-, part-, or scene-level semantics (Fig.~\ref{fig:teaser} - segments), while SVG vectorization captures detailed structure within each region (Fig.~\ref{fig:teaser} - vectorized regions), building a hierarchical representation. Unlike prior works, our stroke generation is structured by scene $\rightarrow$ segment $\rightarrow$ SVG region $\rightarrow$ stroke, leading to geometric and semantic alignment.

\section{Method}

We present a region-guided approach for constructing, sequencing, and rendering brush strokes to emulate a pragmatic painting process. 

Given an input image \(X\) and an empty canvas \(C\), our method generates vectorized stroke sequences that align with semantic scene structures. We operate in an unsupervised setting and support diverse image types and resolutions. 

\subsection{Region Extraction Module}

We decompose the image into painting regions via two steps:
\paragraph{Segmentation:} At the segmentation stage, we employ the Segment Anything Model (SAM) \cite{kirillov2023segment} to decompose the input image into semantically meaningful regions. 
Mainly, we utilize the automatic masks generator (AMG) from SAM to generate a series of scene segments. However, we observe that SAM produces overlapping segments with AMG. To refine such segments, we sort masks based on area (low to high) and use intersection over union (IOU) to filter and adjust masks. This process ensures the generation of non-overlapping segments 
\begin{equation}
    Seg_n = SAM (X), \quad where \quad n = 0, 1, 2, \dots, N.
\end{equation}

Here $Seg_n$ denotes the non-overlapping segment obtained from SAM with sorting based on area (high to low). 

\paragraph{SVG Vectorization:}
Upon extracting scene segments using SAM, we further process each segment to decompose it into vectorized regions through SVG vectorization. For a given segment ($Seg_n$), which is a raster image, the vectorization method approximates vector curves based on the underlying pixel data. This process generates vectorized regions, often referred to as {\Bezier} patches, characterized by curves and fills.

\begin{equation}
    \begin{aligned}
        V\_Region_m &= \text{Vect}(Seg_n), \\
       \text{where} \quad m &= 1, 2, \dots, M, \quad n = 1, 2, \dots, N.
    \end{aligned}
\end{equation}

Here $V\_Region_m$ denotes the vectorized region extracted from $Seg_n$. These vectorized regions not only build an intrinsic hierarchical representation of the segments but also aid in stroke composition by providing insights into the composition of the input scene, linking geometric features with underlying visual characteristics. 
Typically, vectorized regions, referred to as {\Bezier} patches, are composed of various vector curves, including Line, Quadratic {\Bezier} Curve (QBC), Cubic {\Bezier} Curve (CBC), Circular Arc (CA), and Elliptical Arc (EA). The number of control points required to define each curve is as follows: 2 for a Line, 3 for a QBC, 4 for a CBC, and 3 for both CA and EA. Formally, vector curves are defined as $[x_n, y_n, \#color]$. 

\subsection{Sequential brush stroke construction and rendering module}
\paragraph{Sequencing:} After extracting vectorized regions, we organize them into coherent groups guided by principles of proximity, drawing inspiration from perceptual grouping \cite{kubovy2012century, wagemans2012century} in Gestalt theory. Similarity and proximity, two fundamental Gestalt principles, play a pivotal role in human visual grouping and have been widely studied in the contexts of image segmentation \cite{ren2003learning, wang2017unsupervised}, sketch segmentation \cite{qi2013sketching, sun2012free}, and art evolution \cite{prudviraj2025sketch}. From this, we establish sequencing among the vectorized regions using a proximity-based clustering mechanism ($Seq\_Gen$) \cite{prudviraj2025sketch}, where it explores both hierarchical clustering and the Traveling Salesman Problem (TSP) to determine intra-group and inter-group sequencing effectively.
\begin{equation}
    V\_Region_s = Seq\_Gen(V\_Region_m) \quad s = 1, 2, \dots S.  
\end{equation}
where $V\_Region_s$ represents the sequenced vectorized regions. We translate these vectorized regions into polygons for further processing of brush stroke construction and rendering.
\[Polygon_s \simeq  V\_region_s \quad s = 1, 2, \dots S.\]

\paragraph{Stroke construction and rendering:} Starting with an empty canvas $C_0$, we paint step by step, iteratively superimposing the strokes rendered at each stage. At each step $t$, we compute a set of stroke parameters (${S_t}$) to determine the current stroke set, given the region ($Polygon_t$) to be painted and the current canvas $C_t$. 
For each stroke, we generate its corresponding stroke image sequentially and plot it onto the canvas, producing a time-lapse video of how it may have been painted, and the final rendered image ($X_r$). 
Here, we strive to bring the stroke-based painting as close as possible to the original input image, i.e. $X_r \approx X$, in line with our objective of achieving a region-driven pragmatic painting process. 

Following previous methods \cite{liu2021paint, zou2021stylized, hu2024towards}, we explore a parametric rectangular stroke set defined as \(\{x, y, w, h, \theta, r, g, b\}\), where \((x, y)\) represent the coordinates of the origin, \((w, h)\) denotes the dimensions of the stroke, \(\theta\) specifies the angle of counterclockwise rotation, and \((r, g, b)\) defines the template stroke color. 

Here, we approximate the position and dimensions of the stroke using polygon approximation.
For large SVG patches, i.e., polygons with an area exceeding the threshold \(\delta\), a grid decomposition parameter (\(p_{\text{grid}}\)) is applied to divide the polygon into a set of grids. We divide each polygon into a set of grids \cite{wzorek2021polygon,fernandez2000algorithms} by imposing a grid decomposition parameter (\(p_{\text{grid}}\)) and specifying the number of strokes (\(p_{\text{group}}\)) per region. Both \(p_{\text{grid}}\) and \(p_{\text{group}}\) are hyperparameters, configured based on the size of the polygon. 
Then, we approximate each sub-region (\(p_{\text{group}}\)) as a minimum rotated rectangular polygon. And, \(p_{\text{group}}\) directly corresponds to \(Polygon_t\) for polygons with an area less than or equal to \(\delta\), i.e., $p_{\{\text{group}\}}$ == $Polygon_t$. Further, we estimate $\theta$ based on the orientation of the rotated and translated bounding box. 
Additionally, we assign color values ($r, g, b$) based on the fill information of the vectorized regions.
Given a set of stroke parameters and a brush template image, we first create a polygon image ($base_t$) and a brush stroke image using the stroke parameters ($overlay_t$). 
We then perform blending by applying element-wise multiplication across the individual RGB channels and the transparent alpha channel. 

\small
\begin{equation}
    \begin{aligned}
        C_{\text{result}} &= (C_{\text{base}} \odot C_{\text{overlay}}) \cdot (1 - A_{\text{overlay}}) \\
                          &\quad + C_{\text{base}} \cdot A_{\text{overlay}}, \quad C \in \{R, G, B\} \\
        A_{\text{result}} &= A_{\text{base}} \cdot (1 - A_{\text{overlay}}) + A_{\text{overlay}}
    \end{aligned}
\end{equation}

Where $C_{\text{base}}, C_{\text{overlay}}$ are the color channels and \( A_{\text{base}}, A_{\text{overlay}} \) are the alpha channels of the base and overlay images. Once the stroke patch is composed using the computed stroke parameters, it is then applied at the corresponding patch origin on the intermediate canvas $C_t$. These region-guided strokes effectively construct the stroke sequence and contribute to the development of the $\text{stroke} \to \text{region} \to \text{segment} \to \text{scene}
$ progression. Moreover, the vectorized regions retain essential geometric information such as shape, color, position, and opacity, which are crucial for generating effective strokes and producing cohesive artistic paintings.
\section{Experiments}
\label{sec: experiments}

\begin{figure*}
    \centering
    \includegraphics[width=0.9\textwidth]{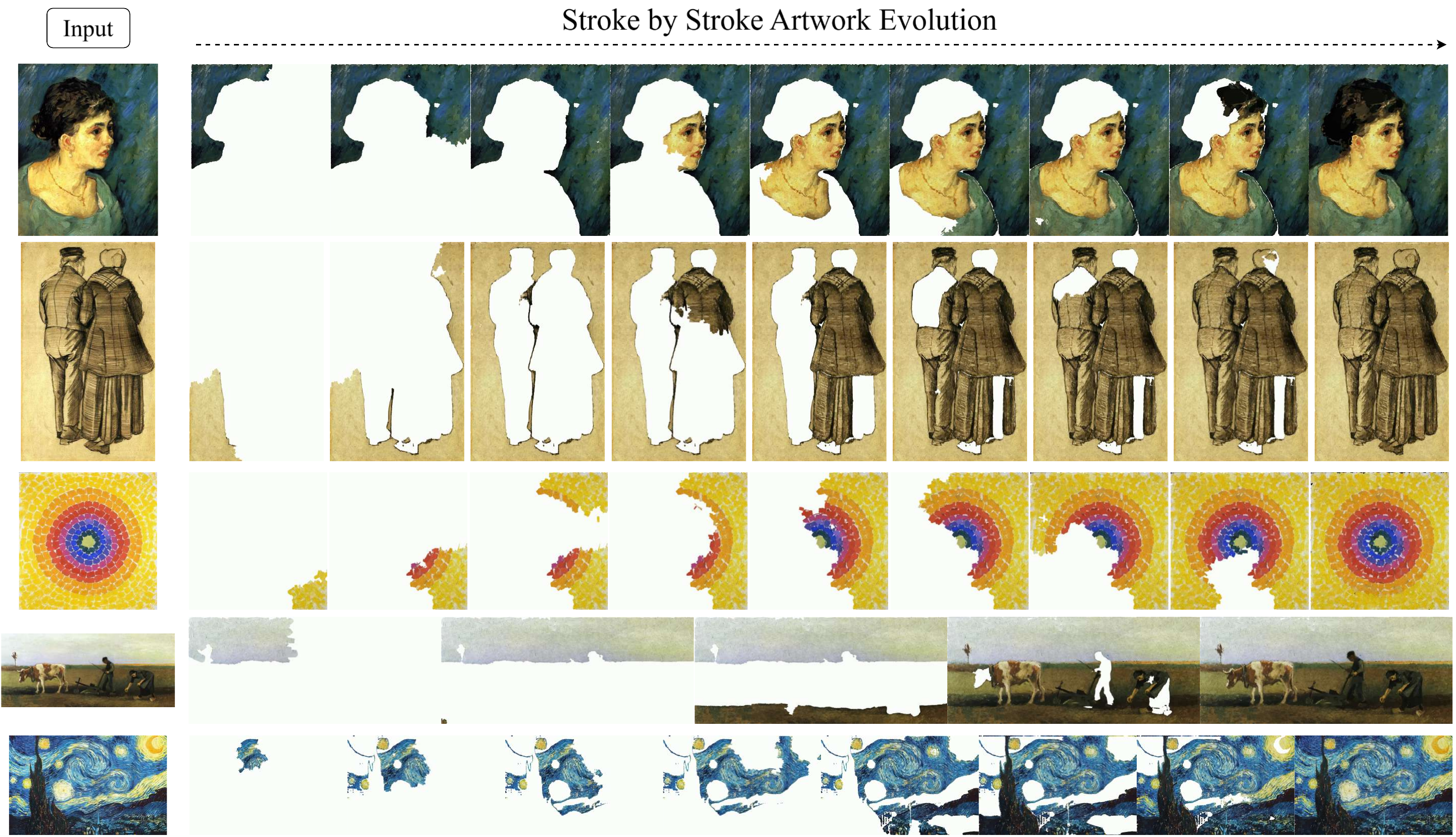}
    \caption{Stroke-by-stroke painting evolution on WikiArt samples. [Zoom in to observe stroke details closely] [The complete painting process across WikiArt inputs is available at \cite{self1}]}
    \label{fig:wikiart}
\end{figure*}

We detail out the data collection, implementation details, and results from the experiments using our methods in the following sections.

\subsection{Dataset}
To evaluate our proposed method across a diverse range of images, we curate samples from the WikiArt dataset \cite{saleh2015large}, METFace \cite{karras2020training}, FFHQ \cite{karras2019style}, IM2Oil \cite{tong2022im2oil}, and Delaunay \cite{gontier2022delaunay}. Specifically,  unlike existing works, we collect images from WikiArt, METFace, and Delaunay to evaluate our model on artworks and demonstrate the artistic painting process on complex paintings. From WikiArt, we sample images of varying resolutions from the `portrait', `landscape/cityscape,' and `unknown genre' categories. Additionally, METFace and FFHQ provide high-quality face paintings and photographic face images, respectively. Meanwhile, Delaunay consists of abstract art images from various artists.

\subsection{Implementation details}
Given an input image, we first exploit SAM to extract non-overlapping segments from the image. Specifically, we employ the ViT-H SAM model with ``points per side'' ranging from 2 to 8, and an IoU threshold and stability score between 0.6 and 0.8. This dynamically determines the number of segments, which varies based on the selected number of points. A higher number of points results in smaller segments, particularly in certain areas with finer details. Once we get segments, we obtain a vectorized image of each segment by tracing the pixel image into vector curves through SVG conversion via a publicly available vectorization tool. Here, we set no constraints on the dimensions of the input image or the number of strokes.
Each vectorized region is further analyzed using the Shapely minimum-rotated triangle \cite{shapely}, which provides positional attributes for the strokes. Based on these attributes, we resize the brush stroke image and rotate it according to the angle of the polygon. Finally, we apply alpha blending between the Shapely polygon and the resized brush stroke image to achieve seamless integration.

\subsection{Results}
Since there is no formal quantitative measure to evaluate the stroke-by-stroke painting process of input images, we primarily assess the effectiveness of our approach qualitatively. Figure \ref{fig:wikiart} illustrates our region-driven painting process on the WikiArt dataset. The stroke-by-stroke evolution includes images from the portrait, unknown genre, and landscape/cityscape categories. From this figure, we observe that the proposed method effectively generates semantic region-based paint stroke sequences for the input image. Moreover, the geometric features derived from the vectorized regions, including the shape, angle, and texture of each stroke, play a key role in enhancing stroke quality and producing a seamless output. 

To verify the adaptability of the proposed method, we further evaluate it on conventional image synthesis datasets such as METFace, FFHQ, Im2Oil, and Delaunay. The stroke-by-stroke painting process on these datasets is demonstrated in Figure \ref{fig:results1}. Specifically, Figure \ref{fig:results1} (a) and (b) show that the generated stroke sequences follow a region-specific semantic order, closely mirroring the natural approach an artist is likely to adopt. As seen in Figure \ref{fig:results1} (c), our proposed method handles complex natural images by effectively interpreting the pragmatic painting process. In Figure \ref{fig:results1} (d), we demonstrate the adaptability of our approach to abstract images. Figures \ref{fig:results1} (e) and (f) illustrate the applicability of our method to the images obtained through generative and style-transfer methods. These results highlight the robustness of our approach in handling images with diverse artistic transformations, preserving the structural integrity of the underlying content.

We evaluate our work against three notable SBR methods: the optimization-based method Stylized Neural Painting (SNP) \cite{zou2021stylized}, and the region-based methods Compositional Neural Painter (CNP) \cite{hu2023stroke}, sketch \& paint \cite{prudviraj2025sketch}, and Segmentation-Based Parametric Painting (SBPP) \cite{de2024segmentation}. A qualitative comparison of our method with these approaches is shown in Figure \ref{fig:qualcompare}. Our method distinguishes itself by supporting varied resolutions and aspect ratios while preserving artwork quality, unlike existing methods \cite{zou2021stylized, hu2023stroke, de2024segmentation} that often lack compatibility with different image dimensions. For instance, SNP is limited to square formats, often resulting in blurry outputs. Both CNP and SBPP generate strokes within grid-based partitions, which restrict the natural semantic flow of region-guided stroke generation. Among these methods \cite{zou2021stylized, hu2023stroke, de2024segmentation}, SBPP is capable of handling high-resolution image inputs and produces satisfactory strokes.  Though these methods generate good results, they produce a limited number of strokes with high computation time. Sketch \& Paint treats the entire painting as a single unit, failing to account for regional grouping and merely overlays SVG patches rather than rendering with brush strokes.
\section{Conclusion}
We introduced a region-driven stroke-based rendering method that integrates image segmentation with SVG vectorization to emulate the step-by-step evolution of artworks. While existing approaches rely on uniform grids or pixel-level optimization, our method organizes strokes hierarchically based on semantic regions, producing structured stroke sequences that better reflect the artistic process. We also demonstrate that our method is effective across artwork and photographic images for high-quality and interpretable stroke-renderings.
\balance
{
\bibliographystyle{ieeenat_fullname}
\bibliography{main}
}

\appendix
\onecolumn
\section{Appendix}
\begin{figure*}[h!]
    \centering
    \includegraphics[width=0.9\textwidth]{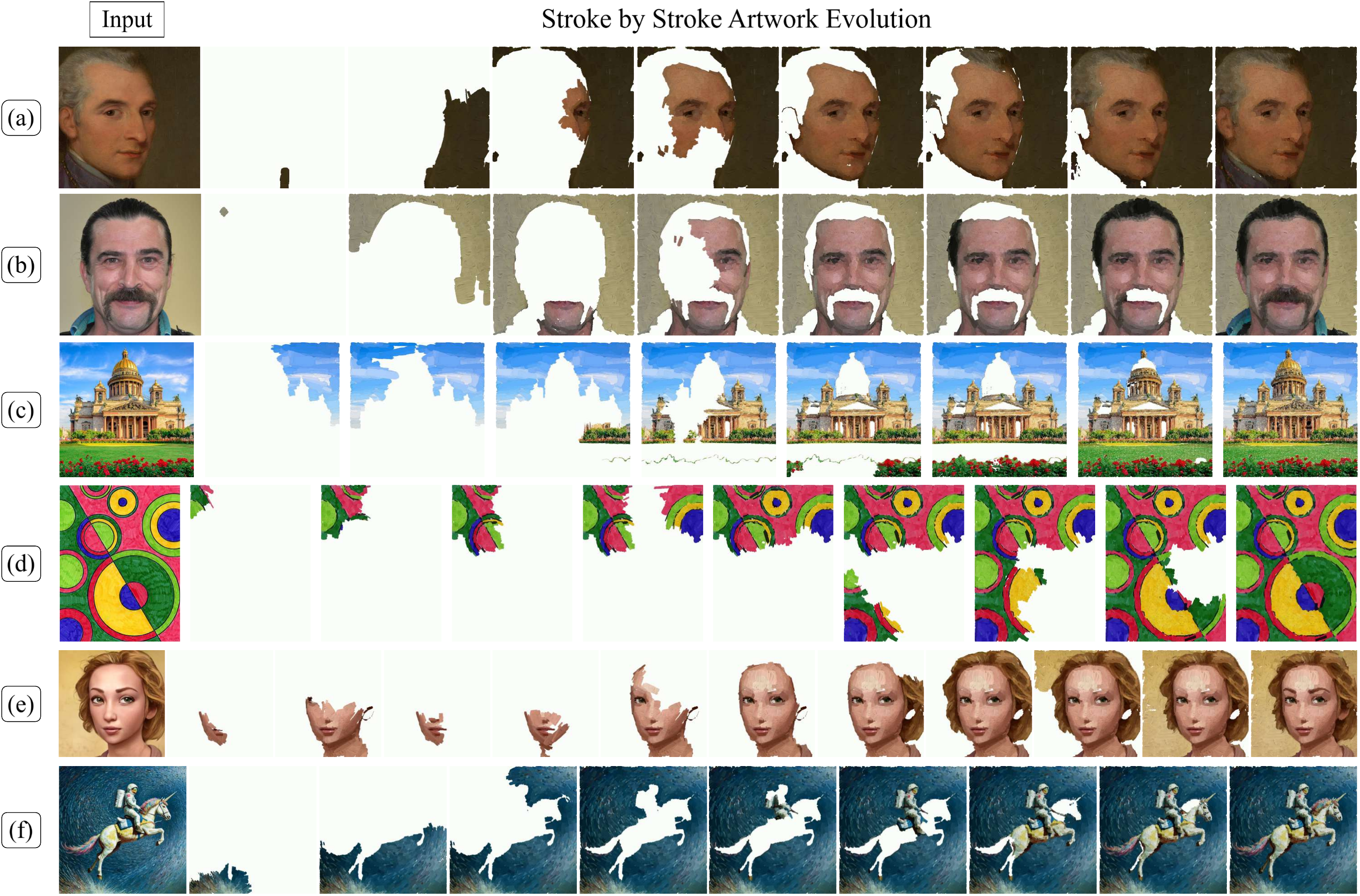}
    \caption{Illustration of Stroke-by-stroke painting evolution on a) METFace \cite{karras2020training} b) FFHQ \cite{karras2019style} c) Im2Oil \cite{tong2022im2oil} d) Delaunay \cite{gontier2022delaunay} e and f) Style transferred images using \cite{yang2022Pastiche} and \cite{replicate_style_transfer}, respectively. [Zoom in to observe stroke details closely] [The complete painting process across is available at \cite{self2}]}
    \label{fig:results1}
\end{figure*}
\begin{figure*}[h!]
    \centering
    \includegraphics[width=0.9\linewidth]{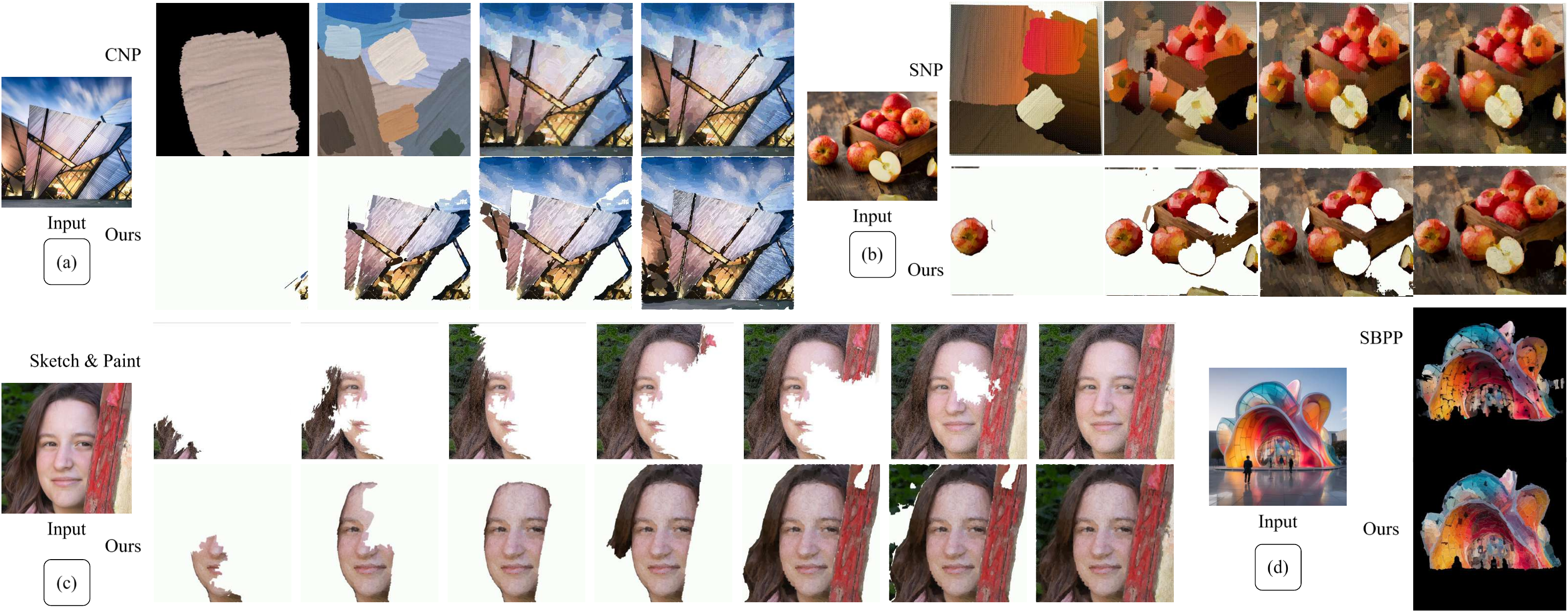}
    \caption{Qualitative comparison of our proposed method with a) Compositional neural painter (CNP) \cite{hu2023stroke} b) Stylized neural painting \cite{zou2021stylized} (SNP) c) Sketch \& paint \cite{prudviraj2025sketch} d) Segmentation-based parametric painting \cite{de2024segmentation}}
    \label{fig:qualcompare}
\end{figure*}

\end{document}